# Training IBM Watson using Automatically Generated Question-Answer Pairs

Jangho Lee
ECE, Seoul National University
ubuntu@snu.ac.kr

Gyuwan Kim
ECE, Seoul National University
kgwmath@snu.ac.kr

Jaeyoon Yoo
ECE, Seoul National University
yjy765@snu.ac.kr

Changwoo Jung
IBM Korea
jungcw@kr.ibm.com

Minseok Kim
IBM Korea
misekim@kr.ibm.com

Sungroh Yoon[*]
ECE, Seoul National University
sryoon@snu.ac.kr

**Abstract**

*IBM Watson is a cognitive computing system capable of question answering in natural languages. It is believed that IBM Watson can understand large corpora and answer relevant questions more effectively than any other question-answering system currently available. To unleash the full power of Watson, however, we need to train its instance with a large number of well-prepared question-answer pairs. Obviously, manually generating such pairs in a large quantity is prohibitively time consuming and significantly limits the efficiency of Watson's training. Recently, a large-scale dataset of over 30 million question-answer pairs was reported. Under the assumption that using such an automatically generated dataset could relieve the burden of manual question-answer generation, we tried to use this dataset to train an instance of Watson and checked the training efficiency and accuracy. According to our experiments, using this auto-generated dataset was effective for training Watson, complementing manually crafted question-answer pairs. To the best of the authors' knowledge, this work is the first attempt to use a large-scale dataset of automatically generated question-answer pairs for training IBM Watson. We anticipate that the insights and lessons obtained from our experiments will be useful for researchers who want to expedite Watson training leveraged by automatically generated question-answer pairs.*

## 1. Introduction

Question answering (QA) is a subfield of natural language processing (NLP) and information retrieval (IR) [1] [2] [3] [4]. The purpose of QA is to find and return a specific and useful piece of information to the user in response to a question [2] [3] [5]. A QA system is a software system designed to answer questions that are posed to in natural languages. IBM Watson is different from the conventional QA systems in that it uses more than 100 different sophisticated techniques for carefully analyzing natural languages [5]. This makes Watson a cognitive computing system that can potentially observe, interpret, and evaluate as humans do [6]. Watson takes a large number of documents and learns question-answer pairs in natural languages when processing questions prepared by the user for training.

Owing to the outstanding NLP processing capability [7] [8], Watson is gradually acquiring a high reputation in the NLP community. IBM Watson is extending its application areas into industry and academia [9] [10] [11]. In the medical industry, Watson Oncology can suggest the best treatment to cancer patients by analyzing clinical information, research material, medical evidence from cancer centers, and the personal information of a patient [10]. Pepper, a robot powered by Watson, brings cognitive computing experiences to everyday lives [11]. To further foster related research in academia, IBM is continuously introducing Watson to universities and recruiting researchers through the Watson University Program available at over 80 universities worldwide [9].

Watson can show its full capability only through sufficient training [12]. More specifically, to train Watson, we need to generate a number of question-answer pairs in natural languages. However, producing a sufficient number of question-answer pairs is usually a labor-intensive task. Lately, machine learning techniques to generate question-answer pairs in large scale have been proposed, and resulting datasets are being released for training large-scale QA systems [13].

---

[*]Corresponding author

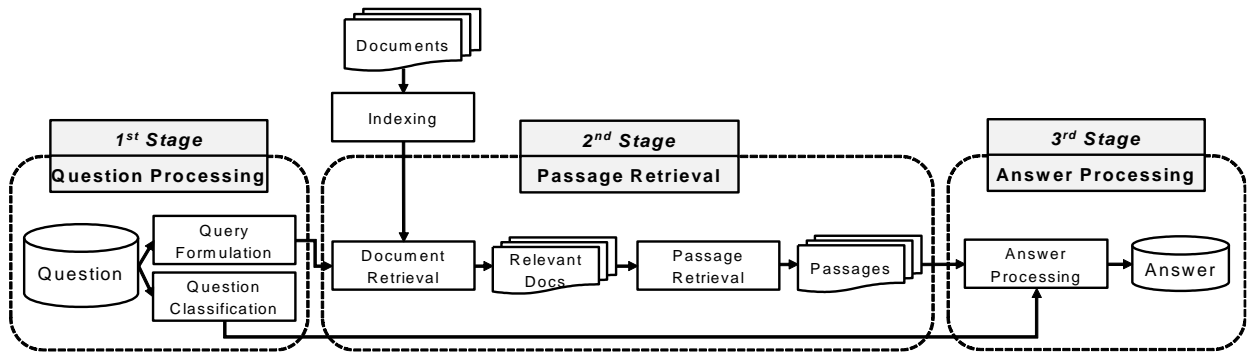

Figure 1. A QA system consists of three stages [5].

These question-answer datasets were generated using deep neural networks [13] and are hopefully expected to reduce the burden of question generation. In this paper, we propose to use such an automatically generated large-scale question-answer dataset for training Watson.

The contributions of this paper can be summarized as follows: (1) We verify the effectiveness of the proposed approach by training Watson using an automatically generated set of question-answer pairs. (2) We propose an automated framework to select questions relevant to QA system training from a large set of question-answer pairs. (3) We demonstrate that training Watson using automatically generated questions along with hand-crafted questions can enhance the overall performance of trained Watson, especially in precision. To the best of the authors' knowledge, our approach is the first attempt to apply automatically generated question-answer pairs to the purpose of training Watson.

The remainder of this paper is organized as follows: Section 2 provides background materials on QA systems and presents related work on Watson research. Section 3 explains the details of our strategy and methods for experiments. Section 4 presents experimental results and discusses the effect of automatically generated question-answer pairs on training Watson and how these data can be used efficiently in large-scale QA systems. Section 5 concludes the paper.

## 2. Background

### 2.1 Review of QA Systems

In QA systems research, we aim to build an automatic system that can retrieve relevant answers when asked questions in a natural language, as most information retrieval systems currently do [2] [3] [14]. A general QA system is composed of three stages as shown in Figure 1 [5]. The first stage of a QA system is to process questions and has two steps: formulation and query classification. In the query formulation step, the QA system extracts queries to get an answer. Next, in the query classification step called *answer type recognition*, the QA system classifies a question according to the expected answer to the question. For example, given the question "Who is the founder of IBM?", we expect an answer type of PERSON. For another question "What is the capital of Republic of Korea?", we expect an answer type of CITY. These tasks are carried out in the question processing stage of the QA system. The second stage is for passage retrieval. In this stage, for each query generated in the previous question processing stage, candidates of the evidence for an answer to the corresponding question are filtered from the passage using the features of named entity information [15], the number of questions, and keywords and n-gram overlaps [16]. The final stage is for answer processing. This process extracts an answer from the result of the second passage retrieval stage. To extract a correct answer, we can use various techniques, such as sentence-pattern matching, answer-type matching, and keyword matching.

### 2.2 Question Categories

There are many ways to categorize questions such as open or closed-domain questions, descriptive questions, and yes/no questions. Descriptive questions include not only definitional questions but also factual questions, which start with an interrogative word, such as *what, where, when, who* and *how* [17]. For example, for the term "IBM Watson," we can generate questions such as "What is IBM Watson?" and "How is IBM Watson used?" Yes/no questions require a statement that indicates whether something is true or false [17]. Examples include "Is there research related to training IBM Watson?"

From the perspective of the types of subjects dealing with questions, a question can be categorized as either open or closed-domain. Open-domain questions consist of diverse topics. The topics of open-domain questions are not limited to a specific area, and training open-domain QA systems thus requires a large amount of data. By contrast, a closed-domain question refers to a question about a specific topic. According to a guide on constructing Watson systems [17], training Watson using specific domain questions helps us establish a baseline system more effectively than using open-domain questions.

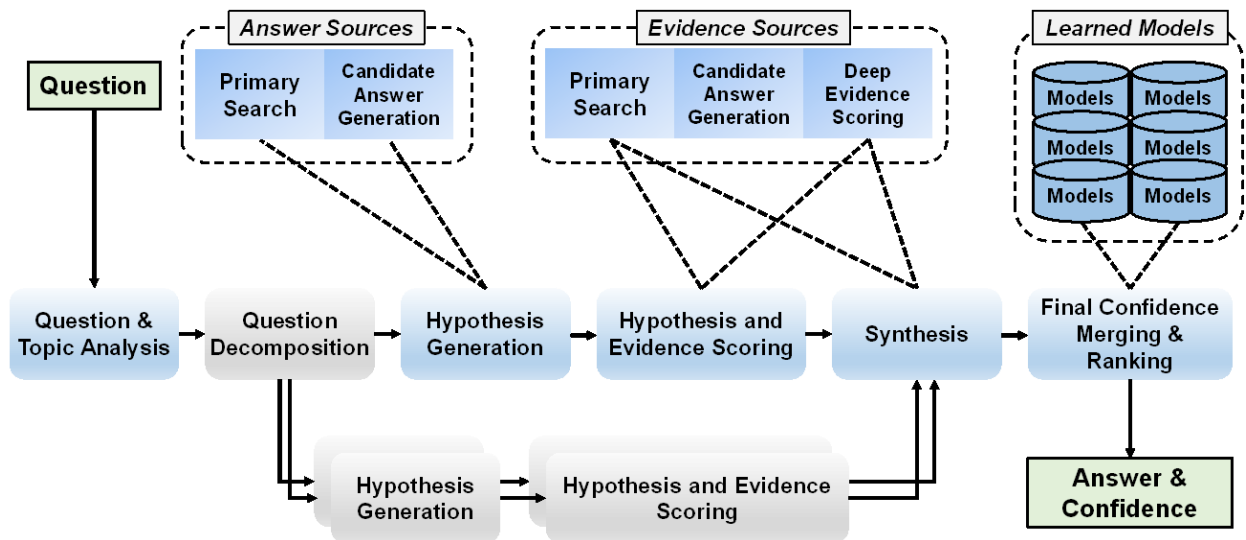

Figure 2. DeepQA architecture [18].

## 2.3 Overview of Watson QA System

Figure 2 illustrates the architecture of the DeepQA [7] technology underlying Watson. DeepQA can find potential answers using NLP techniques such as deep content analysis, information retrieval, and machine learning [7] [8]. In addition, DeepQA is designed to handle a huge volume of data based on big data platforms such as Apache Unstructured Information Management Applications (UIMA) [18] and Apache Hadoop [19]. The QA process starts from constructing a knowledge base[1] which is the search space used as the evidence for Watson to find an answer. After a question enters into Watson, Watson analyzes and decompose it into query languages. Once query languages are extracted, Watson generates hypotheses from query languages and filters out the contents needed to get the correct answer. At the same time, Watson carries out the tasks of collecting evidence, ranking hypotheses, and returning answers that exceed a quality threshold internally defined. Through this whole process, Watson can not only understand questions in natural languages but also answer unseen questions [7]. The biggest difference between traditional QA systems and DeepQA is that the latter is able to extract and accumulate knowledge automatically [7] [20] [21] [22].

Watson Experience Manager (WEM) is a user interface environment that connects a user and a Watson instance [23]. WEM manages the overall processes related to Watson instances. The WEM environment is composed of three parts. The first part is called Manage Corpus which takes and uploads input documents to Watson. Watson can accept various types of documents such as pdf, HTML, XML, and doc. However, not all types of documents are suitable for Watson. Watson prefers documents in well-organized structure such as HTML and XML formats [12] [17]. Watson cannot interpret unstructured data formats, such as diagrams, pictures, and other graphical representations including embedded video, audio, and mathematical expressions [17]. The second part of WEM is for training Watson with prepared training QA pairs as illustrated in Figure 3. To train a Watson instance, the user submits a prepared question to Watson, and Watson then suggests a relevant answer list. In addition, the user can select a paragraph and specify some parts of an input document that can be regarded as an answer. The third part of WEM is for testing. Watson returns an answer paragraph and the associated confidence value in response to a question. A confidence value represents the degree of how much Watson assures that the returned answer is correct [17].

---

[1] A technology used to store and manage complex structured and unstructured information on entities and their interactions used by a computer system (https://en.wikipedia.org/wiki/Knowledge_base).

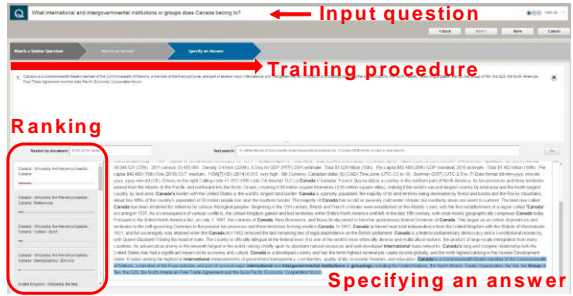

Figure 3. A snapshot of the screen shows the procedure for training Watson.

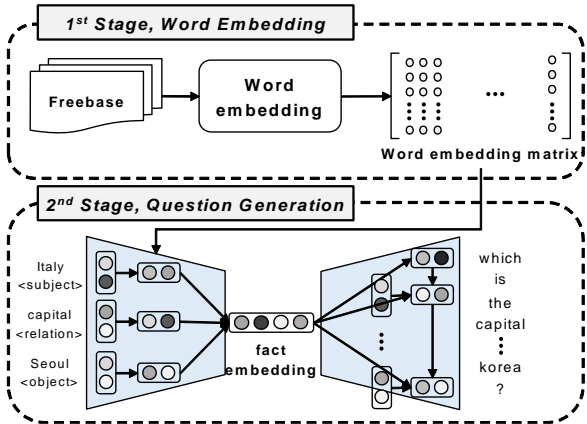

Figure 4. Question-generation model [13].

## 2.4 Related Work

There have been a few studies related to exploring the properties of Watson [12] [23]. As member of the Watson University Program, Murtaza et al. [12] proposed methods and criteria for efficient training of Watson to address the challenge that the internal structure of Watson is like a black box [9]. For example, IBM Watson prefers well-organized texts to a cursory enumeration of sentences from input documents (*i.e.*, unstructured texts). In addition, Murtaza et al. suggested three metrics to evaluate Watson's performance using returned answers and confidence values. These three metrics are recall, accuracy, and precision. Meanwhile, Wollowski [24] reported how to teach the best discipline on using and training Watson in a university class setup. This work provides helpful tips for students to better understand the technical details hidden behind Watson.

In general, to train QA systems, a large number of questions are needed. Research into generating questions has been implemented using diverse strategies [25] [26]. Using deep learning, Serban et al. [13] produced large-scale question-answer pairs and published the resulting data. Although no detail is provided in their paper, we presume that one of the major objectives of this research lied in Watson training. This research is meaningful in that the automatically generated question-answer pairs can be used to relieve at least the burden of making questions. Serban et al. employed Freebase [27] as the source for generating questions. Freebase is an example of a knowledge base consisting of the factual information of entities collected from various sources. (Here, an entity refers to the basic unit for constructing a knowledge base.) The authors generated 30 million question-answer pairs using the entity information contained in Freebase [31]. Figure 4 represents the question-generation module proposed in Serban et al. The first stage is word embedding, which maps natural language words into high-dimensional dense vectors. The second stage represents the encoder-decoder model [13] [28] [29] [30] to generate questions from triplets, each of which consists of a subject entity, an object entity, and relationships between the two entities. The end product of this study was a number of question-answer pairs based on the factual information.

## 3. Proposed Methods

In this section, we provide more details of our methods to prove the effectiveness of automatically generate question-answer pairs [13] [31] for training Watson. The overall procedure consists of three stages, as represented in Figure 5

### 3.1 Stage 1: Data Processing

This stage is to generate a number of question-answer pairs for training Watson and consists of three steps (Steps A, B and C).

In this work, we used an academic version of Watson, which is limited to training with the maximum of 1,000 question-answer pairs [13] [31]. To overcome this limitation, we preprocessed the raw data (the 30 million question-answer pairs [31]) and extracted a subset of question-answer pairs suitable for training Watson. For comparative analysis of multiple training scenarios, we prepared two types of datasets. One was a set of automatically generated questions selected from the 30 million question-answer pairs, and the other was a set of hand-crafted question-answer pairs generated by the researchers participating in this work.

We now elaborate each of the three steps in the first stage of our approach.

**Step A:** The 30 million question-answer pairs were originally produced from the Freebase knowledge base

[27], and these question-answer pairs cover diverse topics. This variety of the topics covered makes them a suitable source for question generation. As far as Watson training is concerned, however, it is known that training Watson for a specific domain is normally easier than training it for multiple domains of various topics simultaneously [17]. For thorough analysis in a controlled setting, we decided to focus on training Watson for a specific domain by using closed-domain questions relevant to the domain.

In order to select the domain suitable for our experiments, we measured the frequency of each entity included in the 30 million QA data and chose the domain that had most frequent usages in the data. The details are as follows. First, we converted the code of a Freebase entity to that of the corresponding Wikidata entity [32], using the mapping file available at the Freebase website. This mapping was needed due to a minor technical issue: The original 30 million QA dataset is based on Freebase, but it is discontinued at the time of writing. For the sake of convenience, we abbreviate Freebase code to Fcode and Wikidata code to Qcode, as shown in Figure 5. Second, we queried the Qcode of a Freebase entity to Wikidata in order to retrieve the name of the entity whose code was queried. Lastly, for each Wikidata entity name obtained as above, we counted the number of its occurrences in the 30 million QA pairs.

As show in the table of statistics in Figure 5A, most of the entities turned out to be related to locations (*e.g.*, USA, Korea, Hawaii, Waikoloa, and Seoul) or personal information (*e.g.*, engineer, actor, and film director). We decided to use the entities related to locations since the entities containing personal information contained mostly low-level details inappropriate for our experiments. We thus selected the entities related to nations and filtered out the 10 most frequent entities. At the completion of Step A, we extracted 1,847,852 questions containing the top 10 frequent entities shown in Table 1.

**Step B:** Although we extracted a number of potential questions in Step A, we had to reduce the number of questions further, as the academic instance of Watson can take questions up to 1,000 for training, as previously mentioned. In this step, we utilized Wikipedia [33] for reduction of QA pairs. As Wikipedia is also a type of knowledge base, we determined to use the entities (*i.e.*, hyperlinked words) available in the Wikipedia pages. Recall that each of the questions extracted from the 30 million question-answer pairs contains two entities, namely a subject entity and an object entity. If a question contained one of the top 10 entities (listed in Table 1) and a hyperlinked word (*i.e.*, a Wikipedia entity with its own pages) together, we selected this question and included it in our training data. The main reason for using hyperlinked words was that we assumed that those entity words would be more relevant to the nation entities rather than random words (non-entity terms without hyperlinks) in the Wikipedia pages.

Through the procedures as outlined above, we reduced the number of questions to 7,060. Out of these, we further selected 400 training and 100 test questions based on the validity of a questions. For example, the question of "Where was country born?" is logically incorrect, and such questions were filtered out.

**Step C:** In addition to extracting 400 training and 100 test questions from the 30 million question-answer pairs, we manually generated questions using the Wikipedia corpus associated with the 10 most frequent entities, as shown in Figure 5C. In order to generate such questions, the six researchers participating in this work cooperated and generated 400 training questions and 100 test questions under the guideline of Murtaza et al. [12].

The details of the overall quantity of data we used in our experiments is listed in Table 2. Specific examples of automatically generated questions (those from Step B) and hand-crafted questions (those from Step C) are given in Table 3.

### 3.2 Stage 2: Watson Training

In this stage, we trained Watson using the training data produced from Stage 1. As our main goal of this study was to prove the validity of using automatically generated questions along with the feasibility of using them as complementary questions for training Watson, we tested three types of training methods: using only the 400 automatically generated question-answer pairs (we call the set of these pairs AQA in Table 4), using only the 400 hand-crafted question-answer pairs (called HQA in Table 4), and using the combination of the two types of pairs (called AQA+HQA in Table 4). We trained three different instances of Watson with the three different methods. To improve the training performance, we provided human feedback for every QA process. That is, for each question, we checked the answers returned by Watson, selected the best answer, and specified the section of the input text containing the expected answer, as shown in Figure 3.

### 3.3 Stage 3: Watson Testing

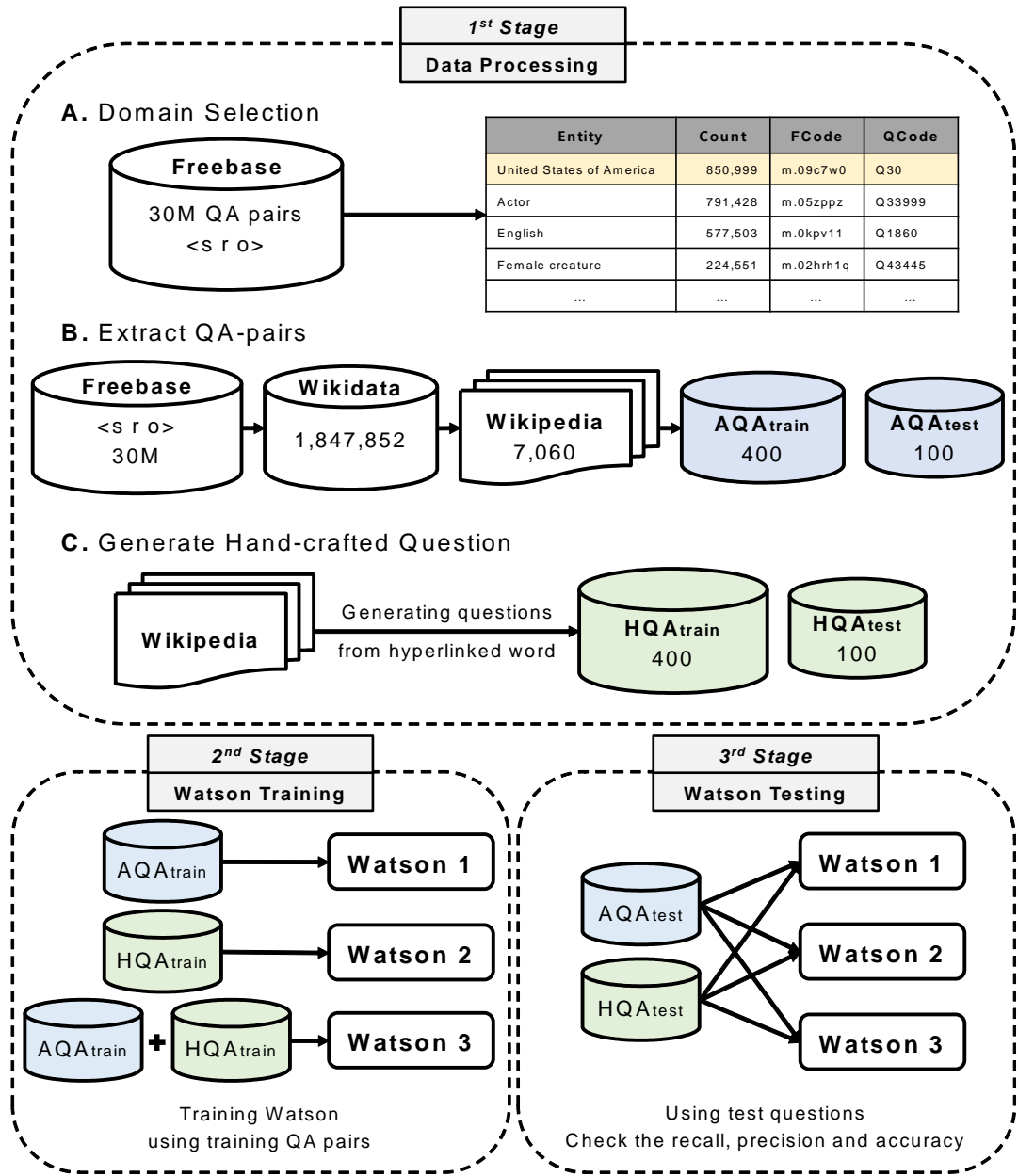

Figure 5. Overview of proposed method.

In this final stage of our approach, we analyzed the performance of Watson trained with the three different methods. To quantify the performance of Watson, we used three metrics (precision, recall, and accuracy) as defined in Murtaza et al. [12]. Note that the definitions of these metrics are different from those used in typical machine learning or information retrieval[2] settings [34]. Adopting the modified metrics was needed since Watson does not directly decide whether an answer is correct or not, but rather returns the paragraph(s) containing an answer along with a confidence value [17].

More specifically, recall is defined as follows. We maintain a counter whose initial value is zero. For each QA pair, we evaluate the answer Watson returns to the question. If the answer is thought to be relevant (including an exact match) to the human answer, then we add 1 to the counter. After evaluating all of the answers, we divide the value of the counter by the number of the total QA pairs used. The result is used as recall. Accuracy is defined similarly to recall, but we increase the counter only when the answer Watson gives exactly matches the human answer. Precision is similar to accuracy, but we increase the counter only if the confidence value returned by Watson with an answer is over 70% (the default value of the confidence threshold used in Watson). Note that, in this setup, recall is always greater than or equal to accuracy, and accuracy is always greater than equal to precision.

---

[2] In information retrieval, precision is the faction of retrieved documents that are relevant to the query; recall is the faction of the documents that are relevant to the query that are successfully retrieved.

## 4. Results and Discussion

Table 4 summarizes our experimental results from training Watson using three different training datasets ($AQA_{train}$, $HQA_{train}$, and $AQA_{train}+HQA_{train}$) and testing each of the trained instances with three different test datasets ($AQA_{test}$, $HQA_{test}$, and $AQA_{test}+HQA_{test}$). Note that there are nine combinations in the training-testing setup, and for each combination, we measured three metric values (precision, recall, and accuracy).

From the experimental results shown in Table 4, we observed that automatically generated question-answer pairs can indeed be used to train instances of Watson, given that the levels of precision, recall, and accuracy were comparable to (or better than) those of training with hand-crafted data. Furthermore, as will be elaborated shortly, we observed that combining automatically generated and hand-crafted data together can boost the performance of Watson in some cases. This suggests that on top of already generated training data for Watson, we can add automatically generated data to gain additional performance boosts.

In our experiments, we often obtained the best results when the types of training and test data match. For instance, when a Watson instance was trained with $AQA_{train}$, testing it with $AQA_{test}$ gave the best results for most cases. On the other hand, when different types of training and test data were used, we observed the degradation of performance, especially in precision. We speculate that the performance of Watson is affected by the question structure used in training and test. The automatically generated question-answer pairs were generated using the Freebase entities, and these questions all have nearly identical structures. On the other hand, the hand-crafted question-answer pairs have more diverse structures. Although the six researchers who participated in generating hand-designed questions tried to follow the same generation rules, it seems inevitable to have personal variations when generating data. In addition, the automatically generated questions are in the form of multiple-choice questions which may have multiple answers, whereas the hand-crafted questions have single answers.

As mentioned earlier, we observed that precision could increase significantly by training Watson using the two types of datasets (AQA and HQA) together. In our experiments, training with two type of questions increased precision approximately 2.6 times higher than training only automatically generated question-answer pairs. Different types of datasets typically have different types of questions, giving different levels of contained information and expressions. As shown in Table 3, the automatically generated question-answer pairs were generated from triplets (subject, object, and their relationship), and their structures are relatively simple. On the other hand, the hand-crafted questions have more diverse structures, and answering them requires various information (not only simple facts but also logical orders and inferences as well). For this reason, we believe that training Watson with two types of question datasets enabled Watson to learn more diverse question patterns than training with a single type. Unlike precision, we have mixed results regarding recall.

According to our definition of performance metrics, there is no trade-off between precision and recall, unlike conventional settings (recall that recall is always greater than or equal to precision in our definition). Nonetheless, we observed that there is an empirical relationship between precision and recall in our experiments. Recall slightly decreased when we trained Watson using the combined dataset compared to training Watson only with one type of question-answer pairs. In other words, increasing precision often resulted in decreased recall, which is compatible with what we normally observe in machine learning or information retrieval. The same line of logic as used for explaining typical precision-recall trade-offs could be used to explain this empirical interplay between precision and recall: Making Watson focus on a narrow search space allows us to have higher precision but negatively affects recall.

Table 1. Top 10 frequent nation entities in Freebase

| No. | Entity | No. | Entity |
|---|---|---|---|
| 1 | United States of America | 6 | France |
| 2 | United Kingdom | 7 | India |
| 3 | Italy | 8 | England |
| 4 | Germany | 9 | Japan |
| 5 | Canada | 10 | Australia |

Table 2. The number of training and test questions

| Question Type | # training question | # test question |
|---|---|---|
| Auto generated (AQA) | 400 | 100 |
| Hand-crafted (HQA) | 400 | 100 |
| Combined | 800 | 200 |

Table 3. Examples of questions used in experiments

| Question Type | Examples |
|---|---|
| Automatically generated | What is the name of a major town in Canada? What is the administrative division of Japan? Which campus is located in Australia? What is a place in Japan? |
| Hand-crafted | Which climate does Hokkaido, Japan, have? What does the term Great Britain refer to? How many companies are listed in the Toronto stock exchange? What is the main reason for the rapid increase in population in Canada? |

Table 4. Summary of experimental results

| Test data | Metric (%) | Training data | | |
|---|---|---|---|---|
| | | $AQA_{train}$ | $HQA_{train}$ | $AQA_{train}$ + $HQA_{train}$ |
| $AQA_{test}$ | Recall | **67** | 18 | 59 |
| | Accuracy | **41** | 7 | 39 |
| | Precision | 27 | 0 | **40** |
| $HQA_{test}$ | Recall | 37 | **46** | 42 |
| | Accuracy | 17 | 27 | **34** |
| | Precision | 2 | 11 | **34** |
| $AQA_{test}$ + $HQA_{test}$ | Recall | **52** | 32 | 50.5 |
| | Accuracy | 29 | 17 | **36.5** |
| | Precision | 14.5 | 5.5 | **37** |

In this work, we used the metrics proposed in Murtaza et al. [12] to assess the effectiveness of Watson training. Although these metrics are reasonably modified over the conventional definitions in order to be used in the context of QA system training, there remains room for improvements. Although changing the internal parameters of Watson would change the accuracy and precision values of an experimental result, these metrics are objective measures in that depending on what Watson returns, the value of accuracy or precision is exactly determined (we directly compare the human answers with Watson answers). On the other hand, the recall metric is a subjective measure affected by who scores the outcome in that measuring relevance may vary from person to person. For this reason, a higher value of recall does not always indicate that Watson can figure out answers in higher quality. To address these issues, we may devise a novel set of metrics that can measure the performance of QA systems including Watson.

## 5. Conclusion

In this paper, we have described our methodology to train IBM Watson using automatically generated question-answer pairs, as an attempt to relieve the burden of manually generating large-scale training data. Through our experiments, we confirmed that our approach is indeed effective for training Watson, delivering competitive performance compared with the conventional training methods. In addition, we demonstrated that training Watson using automatically generated question-answer pairs with hand-crafted question-answer pairs together can allow Watson to provide more accurate answers to unseen questions. Our hope is that the results and insight obtained by this work will help the users of large-scale QA systems make informed decisions on using automatically generated QA pairs for training their systems.

## 6. Acknowledgement

The authors would like to thank IBM Korea for the generous support that made this work possible and Seongsik Park, Jaehong Park, and Jahee Jang at SNU Data Science Lab for their help in question generation. This work was supported by the University Program of IBM Korea. J. Lee, G. Kim, J. Yoo, and S. Yoon were also supported in part by Naver Corp., in part by DataSolution Inc., and in part by SNU ECE Brain Korea 21 Plus project in 2016. This paper was awarded 2016 IBM Best Technology Paper Honorarium.